\documentclass[runningheads, envcountsame, a4paper]{llncs}
\usepackage{amsmath}
\usepackage{amssymb}
\usepackage[pdftex]{graphicx}
\usepackage{epstopdf}
\usepackage[misc]{ifsym}
\usepackage{filecontents}

\usepackage{subfigure}
\usepackage[ruled,vlined]{algorithm2e}
\usepackage{microtype}

\DeclareMathOperator*{\argmin}{arg\,min}

\makeatletter
\def\@fnsymbol#1{\ensuremath{\ifcase#1\or *\or \dagger\or
   \mathsection\or \mathparagraph\or \|\or **\or \dagger\dagger
   \or \ddagger\ddagger \else\@ctrerr\fi}}
\makeatother

\newcommand*\samethanks[1][\value{footnote}]{\footnotemark[#1]}

\begin{document}

\author{Arjun Manoharan\thanks{
        The two authors contributed equally}\inst{1,2} \Letter 
    \and
    Rahul Ramesh\samethanks[1]\thanks{
        Work done primarily while at the Indian
        Institute of Technology Madras}\inst{3} \Letter 
    \and
    Balaraman Ravindran\inst{1,2}}

\institute{Indian Institute of Technology Madras 
    \and Robert Bosch Centre for Data Science and AI \and
University of Pennsylvania \\ 
    \email{arjunmanoharan2811@gmail.com},
    \email{rahulram@seas.upenn.edu},
    \email{ravi@cse.iitm.ac.in}
}

\title{Option Encoder: A Framework for Discovering a Policy Basis in Reinforcement Learning}

\titlerunning{Option Encoder: Discovering a Policy Basis in Reinforcement Learning}
\authorrunning{Arjun Manoharan, Rahul Ramesh and Balaraman Ravindran}

\toctitle{Option Encoder: Discovering a Policy Basis in Reinforcement Learning}
\tocauthor{Arjun Manoharan, Rahul Ramesh and Balaraman Ravindran}

\maketitle              

\begin{abstract}
Option discovery and skill acquisition frameworks are integral to the
functioning of a hierarchically organized Reinforcement learning agent. However,
such techniques often yield a large number of options or skills, which can be
represented succinctly by filtering out any redundant information. Such a
reduction can decrease the required computation while also improving the
performance on a target task. To compress an array of option policies, we
attempt to find a policy basis that accurately captures the set of all options.
In this work, we propose \textit{Option Encoder}, an auto-encoder based
framework with intelligently constrained weights, that helps discover a
collection of basis policies. The policy basis can be used as a proxy for the
original set of skills in a suitable hierarchically organized framework. We
demonstrate the efficacy of our method on a collection of grid-worlds evaluating
the obtained policy basis on downstream tasks and demonstrate qualitative
results on the Deepmind-lab task.

\end{abstract}

\section{Introduction}
Reinforcement learning (RL) \cite{sutton1998reinforcement} deals with solving
sequential decision-making tasks and primarily operates through a
trial-and-error paradigm for learning. The increased interest in Reinforcement
learning can be attributed to the powerful function approximators from Deep
learning. Deep Reinforcement Learning (DRL) has managed to achieve competitive
performances on some challenging high-dimensional domains
\cite{mnih2015humanlevel,mnih2016asynchronous,lillicrap2015continuous,silver2016mastering}.
To scale to larger problems or reduce the training time drastically, one could
attempt to structure the agent in a hierarchical fashion. The agent hence makes
decisions based on abstract state and action spaces, which helps reduce the
complexity of the problem. One popular realization of hierarchies is the options
framework \cite{sutton1999mdps} which formalizes the notion of a temporally
extended sequence of actions.

Discovery of options, particularly in a task agnostic manner often leads to a
large number of options. Option discovery methods
\cite{mcgovern2001automatic,menache2002q,csimcsek2004using,csimcsek2009skill,simsek2005identifying}
as a result, typically resort to heuristics that help prune this set. In such a
scenario, a compression algorithm is of utility, since it would be wasteful to
discard these options and ineffective to use all of them simultaneously. When
using a large number of options, the computation expended for determining the
relevance of each option policy is higher, when compared to using a smaller set
of basis policies \cite{mcgovern2001automatic}. In this work, we demonstrate
that a reduced set of basis policies, results in improved empirical
performances, on a collection of target tasks. 

\begin{figure}[th]
    \centering
    \includegraphics[width=0.82\textwidth]{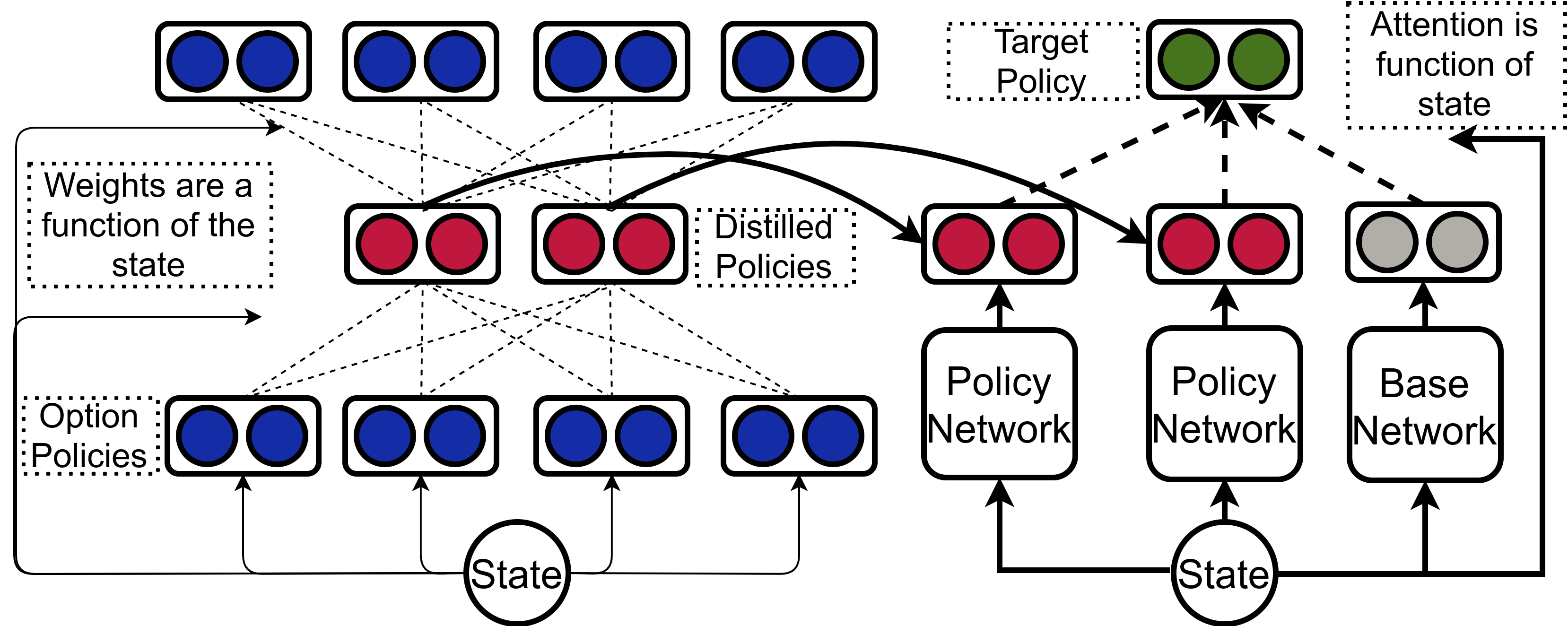}
    \caption{A visual depiction of the Option Encoder Framework. The blue
        colored layers (on the encoder side) correspond to the original option
        policies, and the red layer corresponds to the distilled policies. Any
        set of decoder weights connected to the same output, sum to 1. The
        distilled policies are used in a hierarchical agent that attempts to
        learn a policy for a downstream target task. For more details on the policy
        and the base network in the hierarchical agent, see \cite{rajendran2015attend}.}
    \label{fig:overview}
\end{figure}

Resorting to an existing policy distillation or compression method
\cite{fernando2017pathnet,kirkpatrick2017overcoming,parisotto2015actor} is one
possible alternative. However, these methods distill the options into a single
network, resulting in a single policy that captures the behavior of all
policies. The diversity among the different option policies may be captured
efficiently only if they are distilled to more than one policy. To address the
same, we propose the \textit{Option Encoder}, a framework that attempts to find
a suitable collection of basis policies, from the discovered set of options. We
use an auto-encoder based model where an intermediate hidden layer is
interpreted as a set of basis policies which we term as \textit{distilled
policies}. The distilled policies are forced to reconstruct the original set of
options using attention weights. The intermediate hidden layers would hence be
forced to capture the commonalities between the various options and potentially
eliminate redundancies. The overview of this framework is summarized in
Figure~\ref{fig:overview}. The obtained distilled policies can be used to solve
a new set of tasks and can be used as a proxy for the original set of options in
a new algorithm. 

Our work also provides a simple mechanism to combine options obtained from
different option discovery techniques. This is similar in spirit to
\cite{barreto2019option} but we do not combine the options in a goal-directed
manner. Generating options for a certain goal is useful in some scenarios but is
difficult to work with in a multi-task setting. Using a task-agnostic approach
(like in the Option Encoder) allows the options to be reusable across different
tasks. Furthermore, the Option Encoder discards the full set of expert options
after distilling to a smaller set, which is not the case in
\cite{barreto2019option}. 

Our contributions in this work are as follows: 1) We describe the Option Encoder
framework, which finds a `basis" for a set of options by compressing them into a
smaller set. 2) We present qualitative experiments, analyses and ablation
experiments to justify the efficacy of our framework 3) We empirically
demonstrate that the Option Encoder helps improve or retain the performance on
downstream tasks when compared to using the raw set of options or when using a
reduced number of options. 

The experiments are conducted on a few challenging grid-worlds where we achieve
a 5-fold or 10-fold reduction in the number of options while retaining or even
improving the performance. We also show results in the high-dimensional visual
navigation domain of Deepmind-lab.

\section{Preliminaries}\label{sec:prelim}
RL deals with sequential decision making problems and models the interaction of
an agent with an environment. This interaction is traditionally modeled by a
Markov Decision Process (MDP) \cite{puterman1994markov}, defined by the tuple
$\mathcal{\langle S, A, T}, r, \gamma \rangle$, where $\mathcal{S}$ defines the
set of states, $\mathcal{A}$ the set of actions, $\mathcal{T  : S \times A}
\rightarrow \mathcal{P}(S)$ the transition function (that maps to a probability
distribution over states), $r \mathcal{ : S \times S ' \times A \rightarrow
P}(R) $ the reward function (that maps the current state, next state and action
to a probability distribution over the rewards) and $\gamma$ the discount
factor. In the context of optimal control, the objective is to learn a policy
that maximizes the expected discounted return 
$R_{t} = \sum_{i=t}^{T} \mathbb{E} \left[ \ \gamma^{(i-t)}  r(s_{i}, s_{i+1},
a_{i}) \right]$, where $r(s_{i}, s_{i+1}, a_{i})$ is the reward function. Policy
gradient methods attempt to find a parameterized policy $\pi(a|s;\theta)$ that
maps every state to a probability distribution over the actions, such that the
discounted return is maximized.  Some prominent examples of policy gradient
methods include Advantage-actor critic (A2C) \cite{konda2000actor} and Proximal
Policy Optimization (PPO) \cite{schulman2017proximal}.

\paragraph{\textbf{Options:}} An option \cite{sutton1999mdps} formalizes the
notion of a temporally extended sequence of actions and is denoted by the tuple
$\langle \mathcal{I}, \beta, \pi_o \rangle$. $\mathcal{I} \subseteq \mathcal{S}$
denotes the initiation set of the option, $\beta: \mathcal{S} \rightarrow [0,
1]$  is the probability that the option terminates in state $s$ and $\pi_o$ is
the option policy. In this work, we assume that the initiation set is the set of
all states.

\paragraph{\textbf{Attend, Adapt and Transfer:}}
Attend, Adapt and Transfer architecture (A2T) \cite{rajendran2015attend} is a
model for utilizing expert policies (or options) from $N$ different source tasks
in order to tackle a target task. Consider $N$ expert policies, represented by
$\{ K_i(s) \}_{i=1}^N$. Apart from the expert networks (which are fixed
throughout training), A2T has a trainable \textit{base network} represented by
$K_B(s)$, which is used to learn in regions of the state space, where the set of
experts do not suffice. The target policy $K_T(s)$ is given by:
\begin{equation}
    K_T(s) = w_{N+1}(s) K_B(s) + \sum_{i=1}^N w_i(s) K_i(s)
\end{equation}
The set of weights $w_j(s)$ (for $j \in \{1 \cdots N+1\}$) are attention weights
and as a consequence, satisfy the constraint $\sum_{i=1}^{N+1} w_j(s) = 1$.
$K_T(s)$ is a convex combination of $N+1$ policies and is hence also a valid
policy.

In this work we use a modified version of the A2T algorithm, identical to the
version in \cite{eysenbach2018diversity}. Instead of combining the option
policies using the attention weights, the A2T agent instead selects a single
option policy and persists the same for $T$ steps. Every option terminates, $T$
steps after being selected. We refer to the persistence length $T$ as the \textit{termination limit}.

This modification can be understood as a hard-attention variant of the A2T
framework. The hard-attention weights are trained using A2C since the network
is no longer differentiable. The temporal persistence of the modified A2T
algorithm forces the hierarchical agent to exploit the inherent structure
present in the option policies. We also observed an empirical improvement in
performance with the modified A2T variant and hence use the variant in all our
experiments. Henceforth, any reference to A2T refers to the modified version.

\paragraph{\textbf{Actor-Mimic Network:}} Given a set of source tasks, the
Actor-mimic framework \cite{parisotto2015actor} attempts to learn a network that
copies the policies of the various experts. The loss corresponding to task $i$
is given by:
\begin{equation}
     \mathcal{L}^i_{policy} = \sum_{a \in \mathcal{A_i}} \pi_{E_i}(a|s) \log \pi_{AMN}(a|s; \theta)     
\end{equation}

$\pi_{AMN}$ is a parameterized network that is trained using the cross-entropy
loss. The targets are generated from the expert policy $\pi_{E_i}$. In the case
where the expert consists of Q-values, the targets are generated from the
Boltzmann distribution controlled by a temperature parameter $\tau$.
Parisotto et al. \cite{parisotto2015actor} uses an additional feature regression
loss which we omit in this work.

\section{Option Encoder Framework}
The Option Encoder attempts to find a collection of basis policies, that can
accurately characterize a collection of option policies. Let the policy of an
option $j$, be denoted by $\pi_j$. Let the $i^{th}$ policy of the distilled set
(intermediate layer) be represented by $\pi_i^d$. The set of $M$ ``distilled"
policies $\pi^d = \{ \pi_i^d \}_{i=1}^{M}$ are found by minimizing the objective
given in Equation \ref{eq:obj}. 

\begin{equation}\label{eq:obj}
    \pi^{d^*} = \argmin \limits_{\pi^d } \  \min \limits_{W} \sum_{s \in \mathcal{S}} \sum_{j}
    \mathcal{L} \left( \pi_j(s),  \left( \sum_{i} w_{ij}(s) \times \pi_i^d(s) \right) \right)  
\end{equation}

$W$ is a weight matrix with the entry in $i^{th}$ row and $j^{th}$ column
denoted by $w_{ij}$. The matrix $W$ is such that $\sum_i w_{ij} = 1 \ \forall
j$, which implies that the rows of the matrix are attention weights. $W$ can
also be a function of the state $s$. Each $w_{ij}$ is a scalar such that $ 0
\leq w_{ij} \leq 1$ and it indicates the contribution of distilled policy
$\pi_i^d$, to the reconstruction of option policy $\pi_j$. The function
$\mathcal{L}$ is a distance measure between the two probability distributions
(for example Kullback-Leibler divergence or Huber Loss). The objective states
that a convex combination of the distilled policies should be capable of
reconstructing each of the original set of options, as accurately as possible. 

Equation \ref{eq:obj} is realized using an auto-encoder. The encoder is any
suitable neural network architecture that outputs $M$ different policies. For
example, an encoder in a task with a discrete action space will consist of $M$
different softmax outputs, each of size $|\mathcal{A}|$ (size of action space).
A continuous action space problem will contain $M$ sets of policy parameters
(for example, the mean and variance of a Gaussian distribution). Since the
distilled policies are combined linearly, a single layer for the decoder should
suffice, since the addition of more layers will not add any more
representational power. The entire procedure is summarized in Algorithm
\ref{alg:algo}.

\subsection{Architectural Constraints in the Option Encoder}\label{sec:const}
The architecture is an auto-encoder with two key constraints (see Figure
\ref{fig:overview}). The first restriction is that each distilled policy has a
single shared weight. Alternately, all actions corresponding to a single policy
have the same shared weight. This ensures that the structure in the action space
of the distilled policies are preserved. The second restriction is that the set
of weights responsible for reconstructing any option policy must sum to 1. These
weights are attention weights and can be agnostic to the current state or be an
arbitrary function of it. 

These restrictions are imposed to respect the objective specified in Equation
\ref{eq:obj} i.e., the re-constructed expert policies are convex combinations of
the distilled policies. Furthermore the constraints ensure that the distilled
policies are coherent since a heavily parameterized decoder permits the
information to be captured in the decoder weights, as opposed to the distilled
policies.

To illustrate the utility of these restrictions, consider a scenario in which
all the option policies indicate that the action $a$ has the highest preference
in state $s$. Let action $a$ be assigned the least probability after passing the
options through an encoder. If the weights are allowed to take arbitrary values
on the decoder side, the distilled policies are capable of reconstructing the
options by assigning higher weights to action $a$, even though it has a low
probability as per the distilled policies. Alternately, the decoder can make use
of negative weights to flip the preference order over the actions dictated by
the distilled policies.

\begin{algorithm}[!ht]
\SetAlgoLined
$L$ = Number of rollouts for building distilled policies \;
$N$ = Number of Option policies \;
$M$ = Number of Distilled Policies \;
$K$ = Number of Target Goals \;
$T$ = Number of Steps, an option is persisted \;
$E$ = Number of Episodes for the transfer stage \;
Dataset = Empty list \;

\For{j in (1..... $N$)}{
    env.reset() \;
    \For{i in (1..... $L$)}{
        Get option policies $(\pi_1(s), \pi_2(s), \cdots \pi_N(s))$  \;
        Add $\left(s, (\pi_1(s), \pi_2(s), \cdots \pi_N(s))\right)$ to Dataset\; 
        $a$ = Sample($\pi_j(s)$) \;
        $s$ = env.step($a$)  \;
    }
}

DistillPolicies = train\_auto-encoder(Dataset) \;

\For{j in (1..... $M$)}{
    DistillDataset = None \;
    \For{s in Dataset[0, :]}{
        $\hat \pi_j(s)$ = DistillPolicies(j, s) \;
        add $\hat \pi_j(s)$ to DistillDataset \;
    }
    $\pi_j$ = ActorMimic(DistillDataset) \;
}

\For{i in (1....$K$)}{
    \For{j in (1....$E$)}{
    env.reset() \;
     \While{not done}{ 
             Option\_id = AttentionNetwork.Sample(s) \;
        \For{t in(1...$T$)}{
              a = Option\_id.Sample(s) \;
            s = env.step(a) \;
            store\_transitions()\;
        }
        collect\_rollout() \;
        UpdateAttentionNetwork(rollout) \;
        UpdateBaseNetwork(rollout)
     }
     }
    }
\caption{Summary of the Option Encoder Framework}
\label{alg:algo}
\end{algorithm}

Ideally, one would want the distilled policies to capture the fact that action
$a$ is preferred in-state $s$ among all options. Hence, the proposed two
restrictions ensure that this intended behavior is achieved. The second
restriction also ensures that the output of the decoders are also valid policies
since a weighted combination of the policies (with the weights summing to 1)
will also result in a policy. 

\subsection{Extracting the Distilled Policies}
The current setup would require the execution of the encoder in order to obtain
the distilled policies. This would, however, defeat the entire purpose of the
distillation procedure since the encoder would require the option policies as
inputs. An ideal scenario would allow us to discard the options after the
distillation. In order to achieve the same, the distilled policies (outputs of
the encoder) are utilized as targets to train a network, using an algorithm like
Actor-mimic \cite{parisotto2015actor}. As a result, each distilled policy is
transferred to a network using a supervised learning procedure. The
distilled policies can now be computed from the state, without computing the
option policies. Hence, the network can be used for decision-time planning or
for policy execution in the absence of the original set of options. 

We do not make use of \cite{kirkpatrick2017overcoming,fernando2017pathnet} for
distillation because they are computationally expensive and primarily address
the incremental learning setup. Elastic weight consolidation based training
\cite{kirkpatrick2017overcoming} does not easily extend to multiple tasks and
requires computing the Fisher information matrix which can be expensive for
large networks. Pathnet \cite{fernando2017pathnet} uses a genetic algorithm to
obtain a distilled network, which can be inefficient with respect to the
required number of training iterations.

\section{Experiments}
In this section, we describe experiments designed to answer the following questions:

\begin{itemize}
    \item How do the distilled policies compare against the option policies on a set of tasks?
    \item Why are certain restrictions imposed on the architecture?
    \item Is the performance gain solely due to a reduced number of policies?
    \item Does varying the number of distilled policies affect the performance?
    \item How does the termination limit of the hierarchical agent impact the performance?
\end{itemize}

\subsection{Task description}\label{sec:desc}
\begin{figure}[htb]
    \centering
    \includegraphics[height=0.17\textwidth]{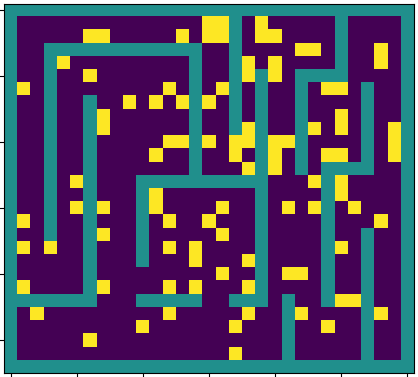} \hspace{0.05cm}
    \includegraphics[height=0.17\textwidth]{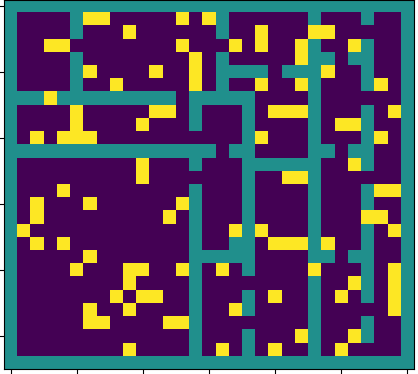} \hspace{0.05cm}
    \includegraphics[height=0.17\textwidth]{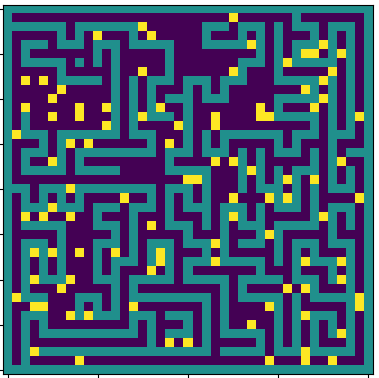}
    \caption{3 grid-worlds are tackled in this work. The yellow dots indicate
    the goals for a collection of tasks that we attempt to solve for in the
grids GW1,GW2,GW3 (left to right)}
    \label{fig:grids}
\end{figure}

\paragraph{Grid-world:} We consider the grid-worlds depicted in Figure
\ref{fig:grids}. The grid-worlds are stochastic where the agent moves in the
intended direction with probability 0.8 and takes a random action (uniform
probability) otherwise. The environment has 4 actions available from every
state, which are up, down, left and right. Each episode terminates after 3000
environment steps. We consider a task where the agent obtains a reward of +1 on
reaching the designated goal and a reward of 0 for every other transition. Fifty
options were learned using the Eigen-options framework
\cite{machado2017laplacian} for each grid-world which were then used to solve
the task of reaching 100 randomly selected goals. These goals are denoted by the
yellow dots on the grid-world in Figure \ref{fig:grids}. Three different
grid-worlds GW1, GW2, and GW3 (left to right in Figure \ref{fig:grids}) were
considered. GW1 and GW2 are of sizes 28x31 each and GW3 is of size 41x41. 

\paragraph{Deepmind Lab:} The Deepmind-Lab domain \cite{beattie2016deepmind} is
a visual maze navigation task where the inputs are images. In this work, the
images are converted to grayscale images before being used as inputs to a
network. The action space is discretized into 4 actions which are forward,
backward, rotate-left and rotate right. Every step receives a reward of -0.01
and the episode ends after reaching a designated goal or after 3000 environment
steps. The agent additionally receives a reward of 1.0 on reaching the goal and
receives a reward of -1.0 if it fails to reach the goal after 3000 steps in the
environment.

\subsection{Architecture overview}

The state in all grid-world experiments is represented as an image of the grid
(with 1 channel) with all zeros, except at the location of the agent. We impose
the encoder to also have shared attention weights (each policy has a single
attention weight) like the decoder. This implies that the original set of
options are combined using attention weights to yield the distilled policies
which are then combined using another set of attention weights to yield the
reconstructions.

\textit{Option Encoder:} The encoder and the decoder are comprised of attention
weights which are functions of the current state. The state-based attention
network consists of two convolution layers (5x5x4 and stride 2 and 3x3x8 and
stride 1) and a fully connected layer with 32 units which then outputs the
attention weights. For the Deepmind-Lab task, the current state is converted
into attention weights using a network that contains 3 convolution layers
(8x8x32 and stride 4, 4x4x64 and stride 2, 3x3x64 and stride 1) followed by a
fully connected layer of size 512 which then outputs the attention weights. 

\textit{Hierarchical Agent:} The A2C algorithm with the modified A2T framework
(described in Section \ref{sec:prelim}) was used to train the agent (referred to
as the A2T + A2C agent). The base network consists of 2 convolution layers (same
configuration as earlier) followed by a fully connected layer of size 128 which
yields the policy and the value function heads. The base network policy and the
option policies are combined using attention weights to yield the final policy.
The termination limit is 20 in this case.

\subsection{Evaluating on Grid-worlds}
\begin{figure}[!ht]
    \centering
    \includegraphics[width=0.37\textwidth]{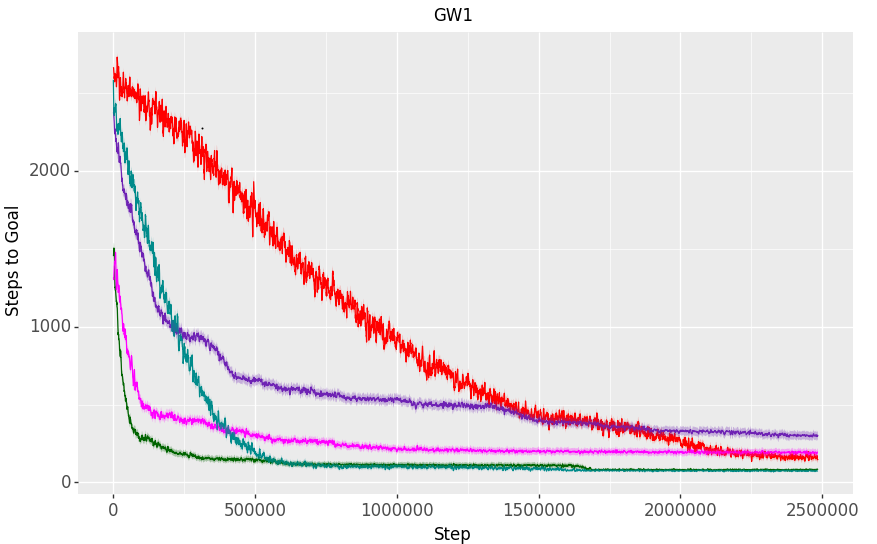}
    \includegraphics[width=0.37\textwidth]{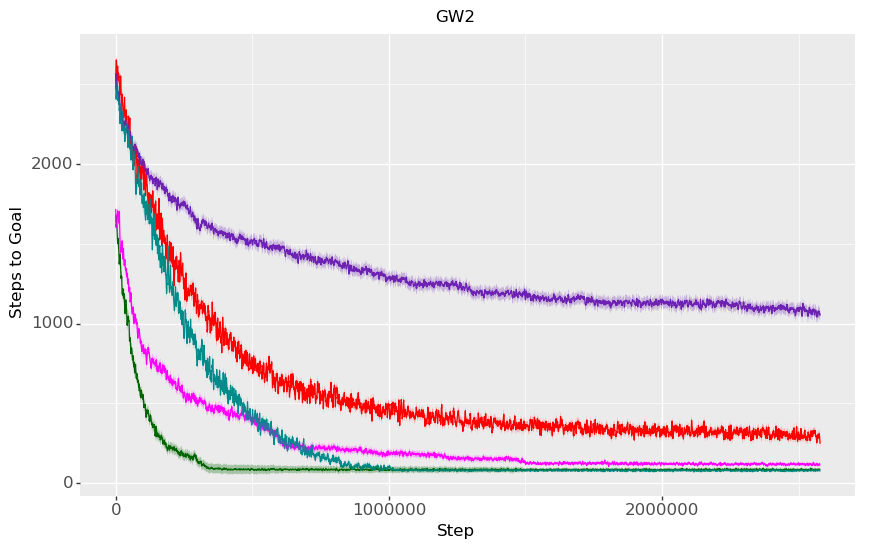} 
    \includegraphics[width=0.37\textwidth]{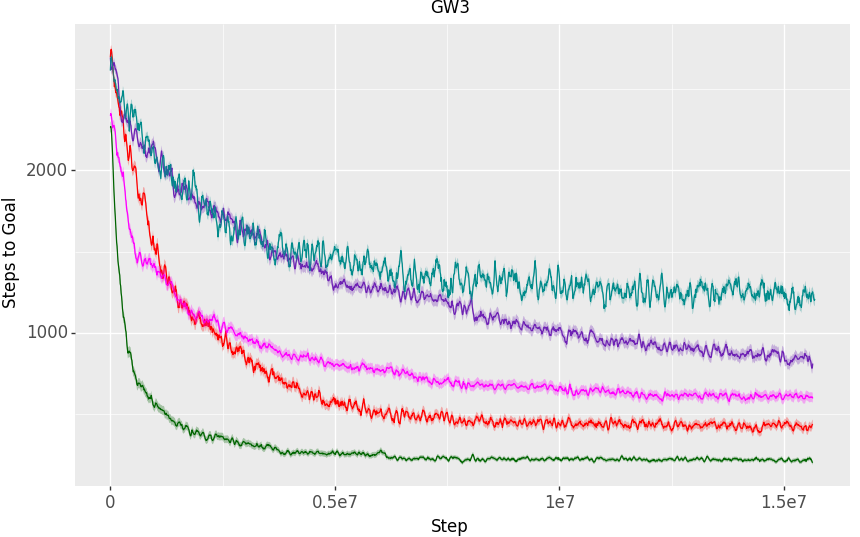} \hspace{0.05cm}
    \includegraphics[width=0.18\textwidth]{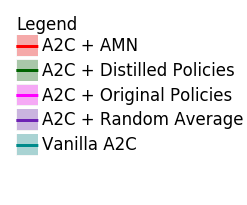}
    \caption{Plot depicting the number of steps to reach the goal (performance
    measure) vs. the number of environment steps on GW1, GW2, GW3 respectively
(left to right)}
   \label{fig:eval1}
\end{figure}

\begin{figure}[!ht]
    \centering
    \subfigure[Distilling varying number of experts to 5 options]{\label{fig:distill5}\includegraphics[width=0.41\textwidth]{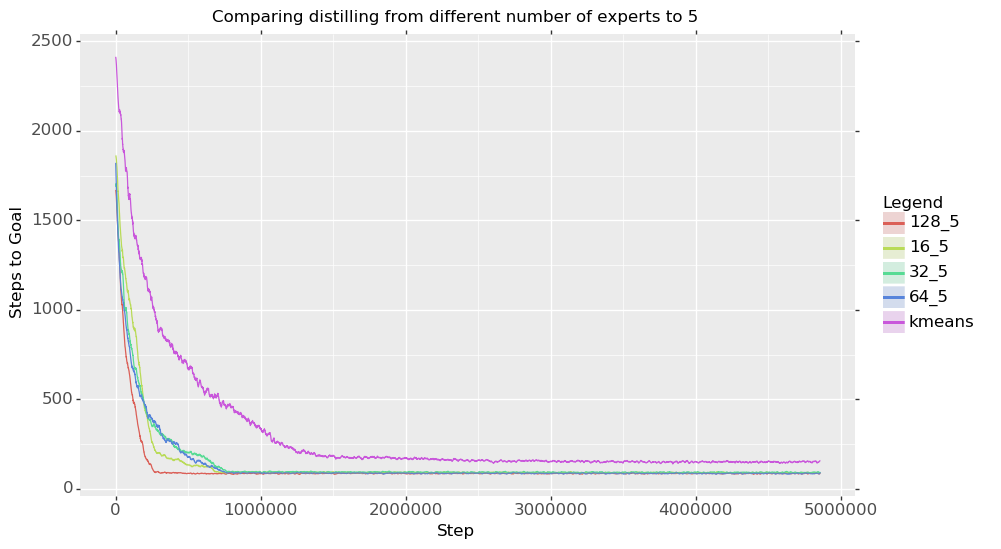}}
    \subfigure[Comparing 10 Eigen-options to 10 options obtained from distilling 64 Eigen-options]{\label{fig:distill6}\includegraphics[width=0.43\textwidth]{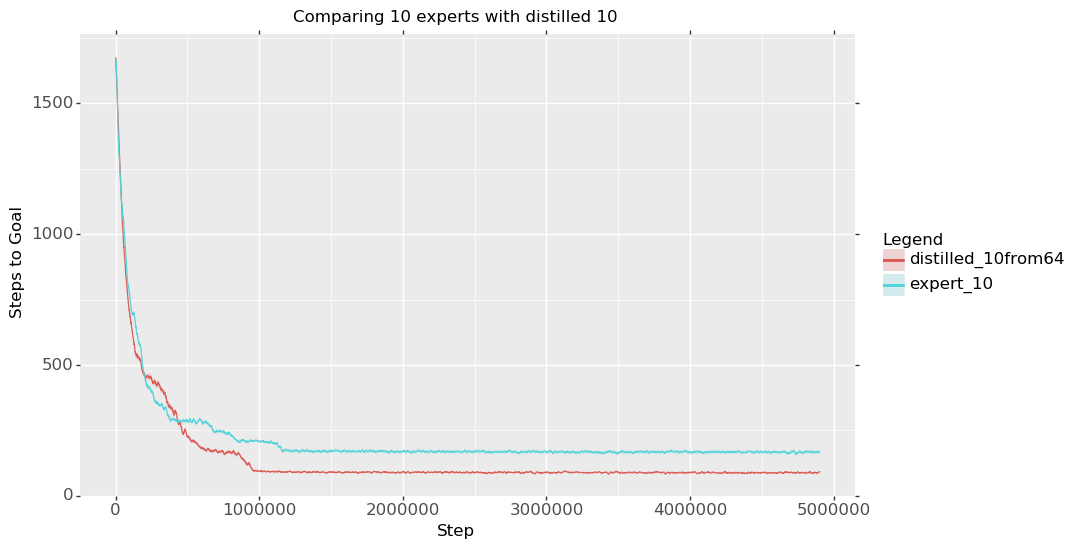}}
    \caption{Option Encoder is a better alternative than discarding the
    discovered options and is capable of reducing a large number of options}
    \label{fig:my_label}
\end{figure}

50 options were obtained using the Eigen-options framework
\cite{machado2017laplacian} where the eigen-vectors of the graph laplacian are
used to define options. The policies corresponding to each eigen-vector are
obtained using a vanilla A2C agent (architecture identical to an A2C+A2T agent
barring the attention). This is followed by the Option Encoder framework, which
distills the option policies obtained from all states in the grid-world, into 5
distilled policies. The A2C+A2T agent can make use of the original set of
options or the distilled set, which we term as A2C + original and A2C +
distilled respectively. For both agents, the selected option is persisted for 20
steps (termination limit) after which a new option is selected. We also evaluate
a vanilla A2C agent. Actor-mimic is another baseline that we consider, where all
the options are distilled to one policy which is used in the A2C + A2T setup (we
refer to this as A2C+AMN). Finally, we consider the random average agent which
consists of an A2T+A2C agent attending to 5 policies and a base network. Each of
the 5 policies are obtained by averaging the policies of 10 randomly chosen
option policies.  

The agents are periodically evaluated every 500 environment steps and the
performance curves are presented in Figure \ref{fig:eval1}. The graphs are
clearly indicative of the fact that the A2C+distilled agent outperforms all
other baselines. Since we tackle 100 different target tasks, the effort required
to obtain the distilled policies (or the options) is negligible when compared to
solving the multi-task problem. Hence, the presented performance curves are
comparable. 

We also vary the number of experts and distill to 5 options (Figure
\ref{fig:distill5}) and notice that an increased number of experts, improves the
performance. This observation is further corroborated by Figure
\ref{fig:distill6} which indicates that the top 10 Eigen-options perform worse,
when compared to using a set of 10 distilled options. In a resource constrained
situation where only few options are required, one can distill knowledge from
many options to a smaller set. We also attempted to cluster the policies using
K-means on the policy space. Unsurprisingly, our distilled policies outperformed the
options obtained using K-means. The centroid of each K-means cluster was used as
a substitute for the distilled policy. We run K-means to discover 5 clusters
from 50 expert policies.

\subsection{Understanding Architecture Constraints}
We enforce certain restrictions on the auto-encoder as described in
Section~\ref{sec:const}. We conduct a qualitative analysis of different
architectural variants. Remember that the Option Encoder architecture requires
the weights to be attention weights and the policy to be a convex combination of
the policies from the previous layers. This implies that all actions of a policy
share a single weight.

We consider the grid-world in Figure~\ref{fig:arch1} with four expert options
going to the 4 corners of the top-left room. Figure~\ref{fig:arch1} represents a
heatmap of the distribution of states visited by the 2 hidden policies generated
from the Option Encoder framework. We visualize the heatmap for following
architectural variants:

\begin{figure}[!ht]
    \centering
    \subfigure[Option Encoder framework]
    {\label{fig:arch1}
    \includegraphics[width=0.18\textwidth]{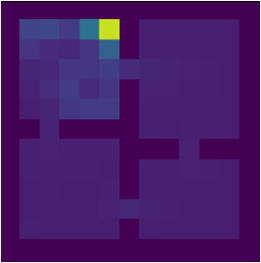}
    \includegraphics[width=0.18\textwidth]{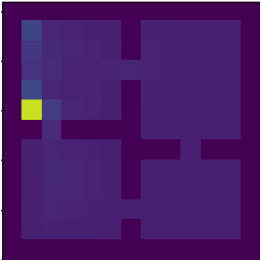}
    \includegraphics[width=0.045\textwidth]{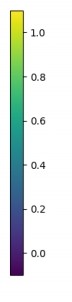}
    }
    \subfigure[No restriction on attention weights]
    {\label{fig:arch2}
    \includegraphics[width=0.18\textwidth]{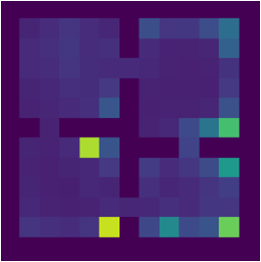}
    \includegraphics[width=0.18\textwidth]{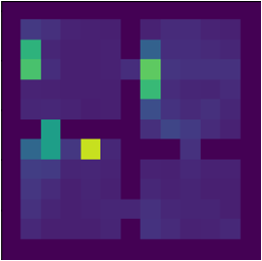}
    \includegraphics[width=0.045\textwidth]{images/arch_legend.jpg}
    }
    \caption{Heatmap of the visitation distribution of two distilled policies}
\end{figure}
\begin{figure}[!ht]
    \centering
    \subfigure[No restriction on weight-sharing for policies]
    {\label{fig:arch3}
    \includegraphics[width=0.18\textwidth]{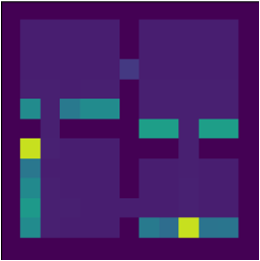}
    \includegraphics[width=0.18\textwidth]{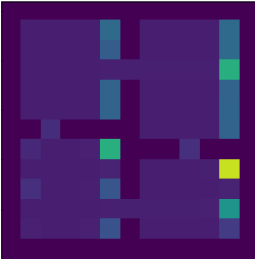}
    \includegraphics[width=0.045\textwidth]{images/arch_legend.jpg}
    }
    \subfigure[No restrictions on both attention and weight-sharing]
    {\label{fig:arch4}
    \includegraphics[width=0.18\textwidth]{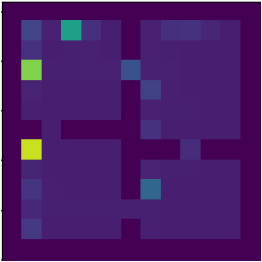}
    \includegraphics[width=0.18\textwidth]{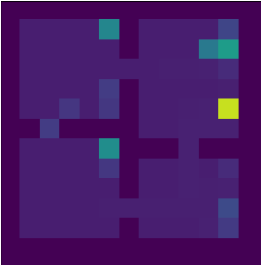}
    \includegraphics[width=0.045\textwidth]{images/arch_legend.jpg}
    }
    \caption{Heatmap of the visitation distribution of two distilled policies}
\end{figure}

\begin{itemize}
    \item \textit{Removing restriction on attention weights}: In this case, the
        weights are not fed through a softmax layer to enforce that they sum to
        1. Figure~\ref{fig:arch2} highlights that we do not learn an
        interpretable policy.
    \item \textit{Removing restriction on weight sharing}: Since the layers
        enforce the output policies to be convex combinations of the input
        policies, each policy is assigned a single attention weight. We instead
        assign a single weight for every action in the policy increasing the
        number of weights by $|\mathcal{A}|$ (size of the action space).
        Figure~\ref{fig:arch3} again highlights that we do no learn
        qualitatively useful policies.  \item \textit{Removing both
        restrictions}: This case corresponds to a vanilla auto-encoder with no
        constraints on the weights. An intermediate layer is interpreted as a
        policy and is found to not be qualitatively interpretable (see
        Figure~\ref{fig:arch4}).
\end{itemize}

\subsection{Understanding the distilled policies}

\begin{figure}[!ht]
    \centering
    \includegraphics[width=0.14\textwidth]{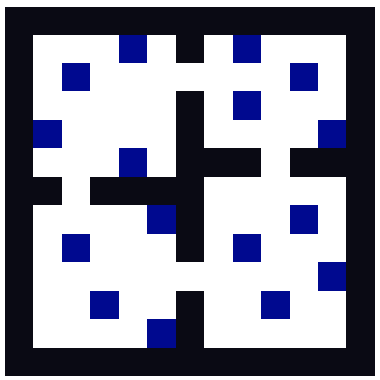} \hspace{0.02cm}
    \includegraphics[width=0.14\textwidth]{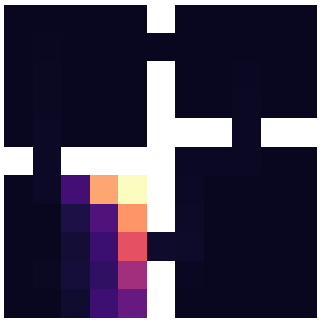} \hspace{0.02cm}
    \includegraphics[width=0.14\textwidth]{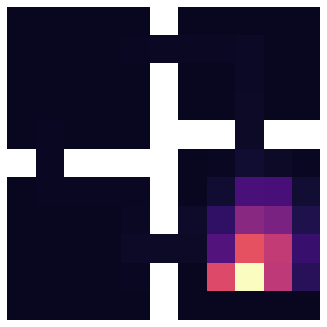} \hspace{0.02cm}
    \includegraphics[width=0.14\textwidth]{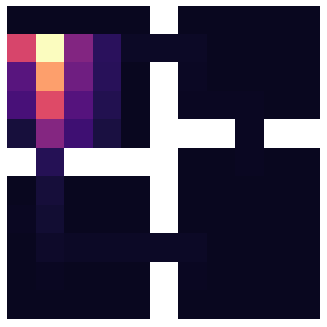}\hspace{0.02cm}
    \includegraphics[width=0.14\textwidth]{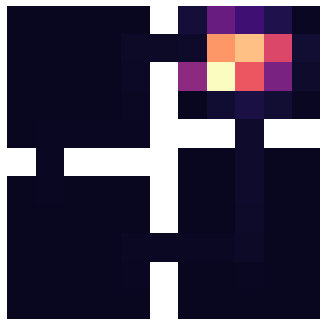} \hspace{0.02cm}
    \includegraphics[width=0.14\textwidth]{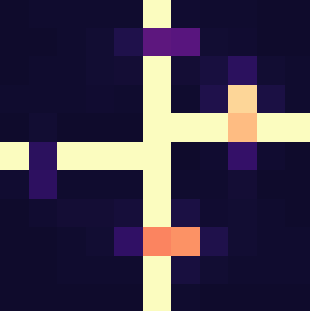}
    \caption{The leftmost figure denotes the set of 16 expert policies while the
    middle 4 figures (2nd to 5th image from left) visualize the visitation count
of the rollouts of the distilled policy. The rightmost figure corresponds to the
visitation count of an Actor mimic network (AMN) trained to distill 16 policies into 1 policy.}
    \label{fig:distill4}
\end{figure}

To understand the Option Encoder framework, we consider 16 expert policies, each
navigating to a specific goal as indicated in the left-most image in Figure
\ref{fig:distill4}. Each blue cell denotes a goal towards which an expert policy
navigates to optimally. These experts are distilled to 4 different policies. In
order to visualize these policies, we develop a heatmap of the visitation counts
for each policy. The heatmaps are obtained by sampling from the respective
distilled policies. The agent is executed for 50,000 steps and is reset to a
random start state after 100 steps in the environment. Figure \ref{fig:distill4}
demonstrates how the Option Encoder framework captures the commonalities between
various policies. We also compare the same with the visitation count plot
obtained from an Actor-mimic network trained by combining all 16 policies. 
\subsection{Randomly sampling a set of options}
\begin{figure}[!ht]
\centering
   \includegraphics[width=0.7\textwidth]{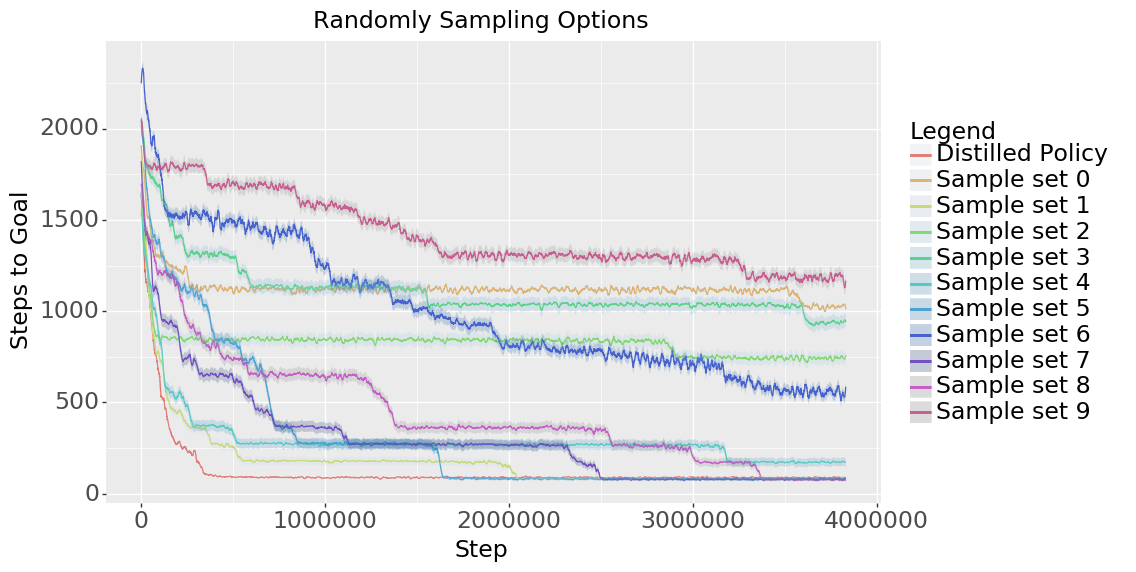} 
   \caption{Performance comparison against a few random subsets of options to
   show that the performance improvement is not due to a reduced number of
   options}\label{fig:sample}
\end{figure}

This analysis on GW2 was conducted to demonstrate that our proposed framework does
not derive a significant advantage from using a reduced number of option
policies. 50 options were divided into 10 sets of options (each of size 5) where each option
appears in exactly one of the ten sets. Figure \ref{fig:sample} shows that a
random sample of options can lead to vastly varying performances (based on the
relevance of the options). However, the distilled policies outperform every set
of random options we consider, hinting at the fact that all the options are
useful for solving a new set of tasks. Each line in Figure \ref{fig:sample}
corresponds to the average performance over 25 randomly selected goals.

\begin{figure}[th]
    \centering
    \subfigure[Performance plotted for a varying number of distilled policies.]
    { \label{fig:distnum1}\includegraphics[width=0.5\textwidth]{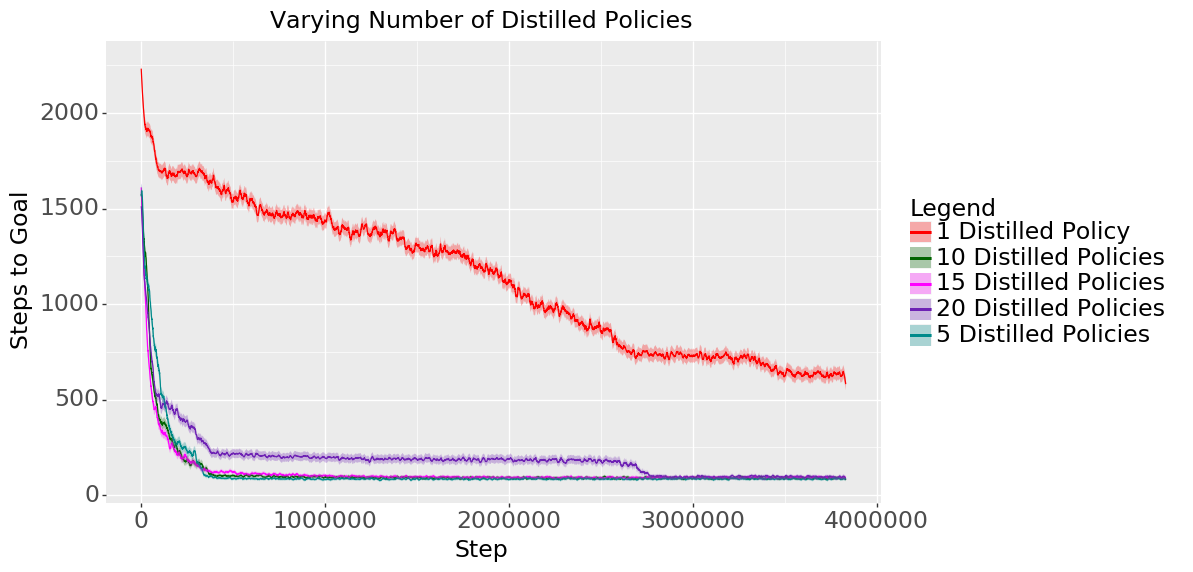}}
    \subfigure[Varying the termination limit of \hspace*{0.55cm}   the A2T + A2C setup]
    {\label{fig:term}\includegraphics[width=0.45\textwidth]{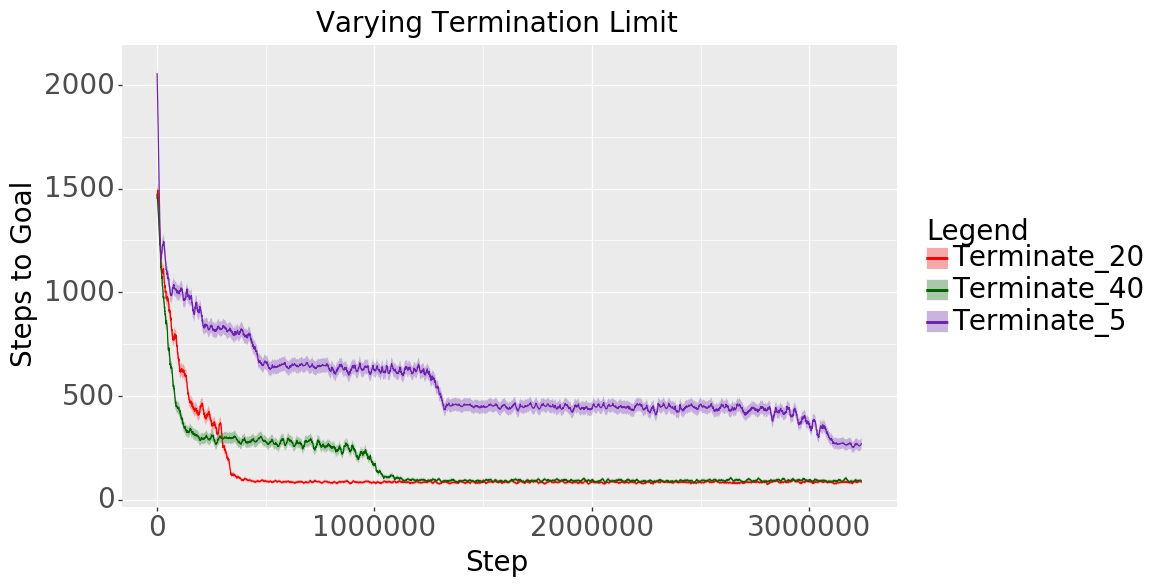}}
    \caption{The Option Encoder is fairly robust to hyper-parameters like the
    number of distilled policies or the termination limit}
\end{figure}

\subsection{Varying the number of Distilled policies}
An experiment on GW2, was conducted (see Figure \ref{fig:distnum1}) to identify
the impact of the number of distilled policies on the final performance. The
plots are indicative of the fact that the agent is not highly sensitive to this
parameter. Distilling to a single policy yields a poor performance curve since a
single policy is incapable of capturing the variety in the fifty expert
policies.

\subsection{Varying termination limit}
We analyze the impact of termination limit $T$ (defined in Section
\ref{sec:prelim}) and obtained the performance curves for GW2. When the
termination limit is low, the agent fails to leverage the knowledge of a useful
sequence of actions since it cycles between the various options. Hence, a higher
termination limit results in a persistent strategy for an extended duration,
thereby yielding better performance (see Figure \ref{fig:term}). However, beyond
a certain value for the termination limit, the performance deteriorates since
the agent spends an excessive amount of time on a single
option, thereby sacrificing some fine-grained control.

\subsection{The Deepmind-lab task}
\begin{figure}[!ht]
    \centering
    \label{fig:dmenv}\includegraphics[width=0.26\textwidth]{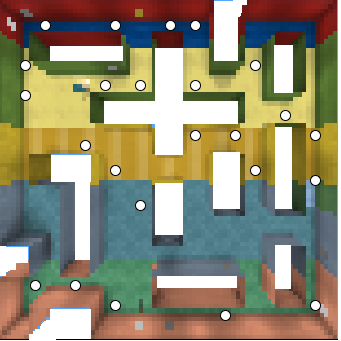}
    \caption{Map used for Deepmind-lab. The white dots are sub-goals for option policies}
\end{figure}

\begin{figure}[!ht]
    \centering
    \hspace{0.8cm}
    \subfigure[8 distilled policies]{\label{fig:dmlab8}\includegraphics[width=0.44\textwidth]{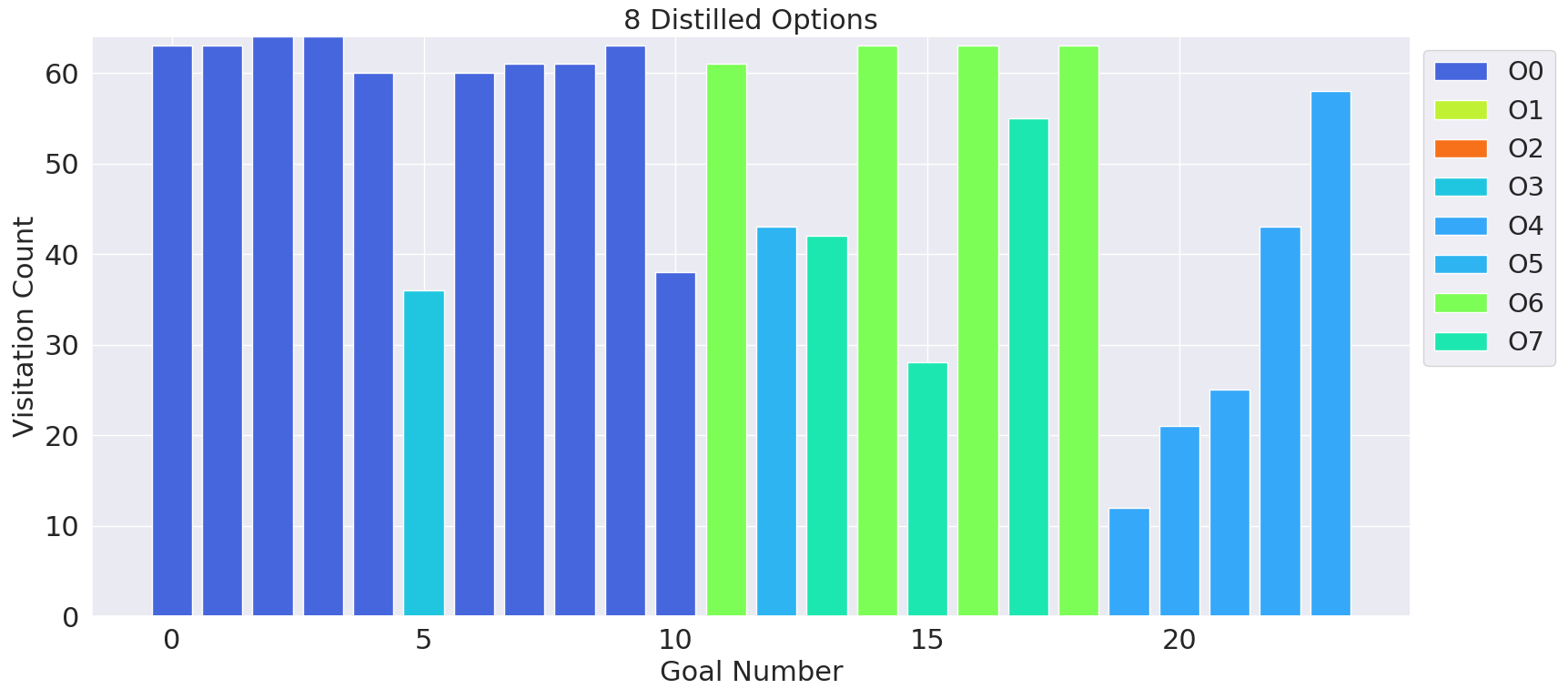}}
    \subfigure[12 distilled policies]{\label{fig:dmlab12}\includegraphics[width=0.44\textwidth]{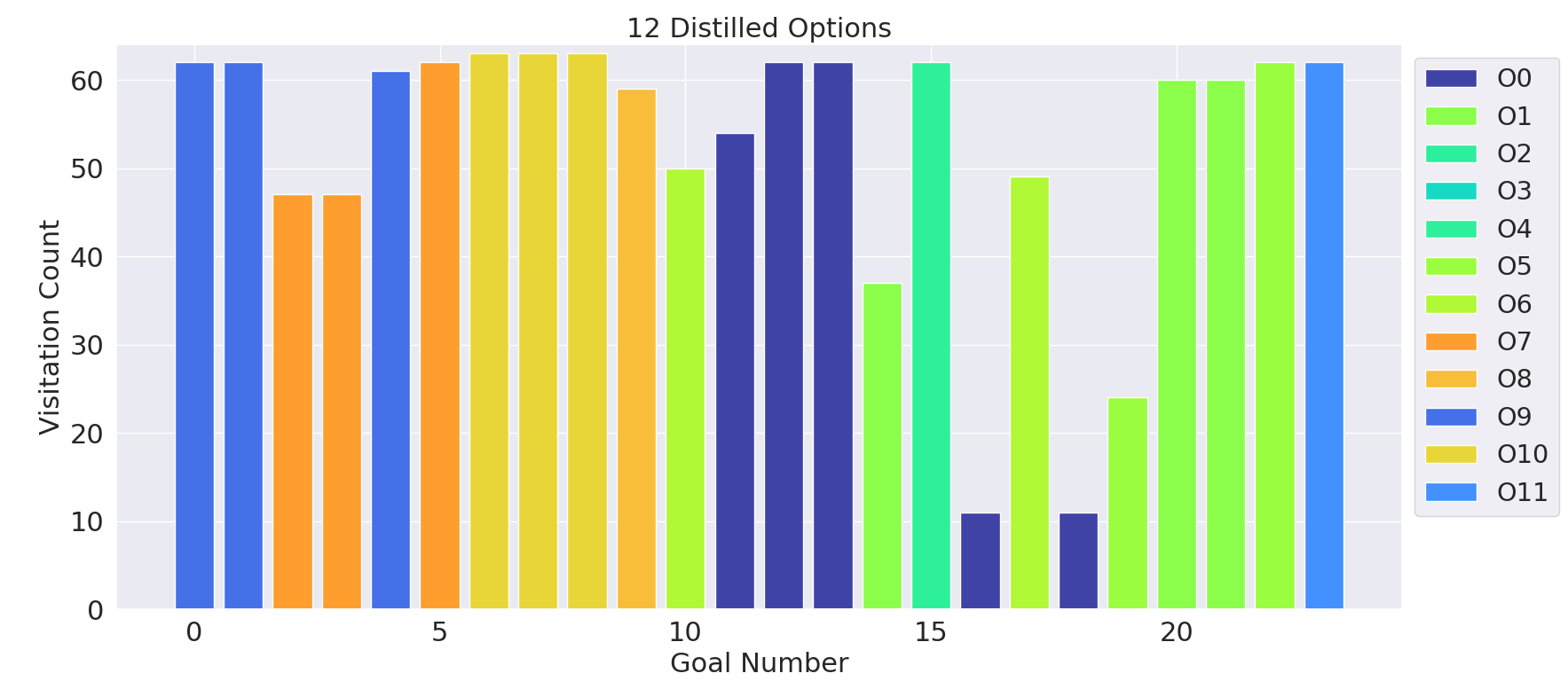}}
    \subfigure[16 distilled policies]{\label{fig:dmlab16}\includegraphics[width=0.44\textwidth]{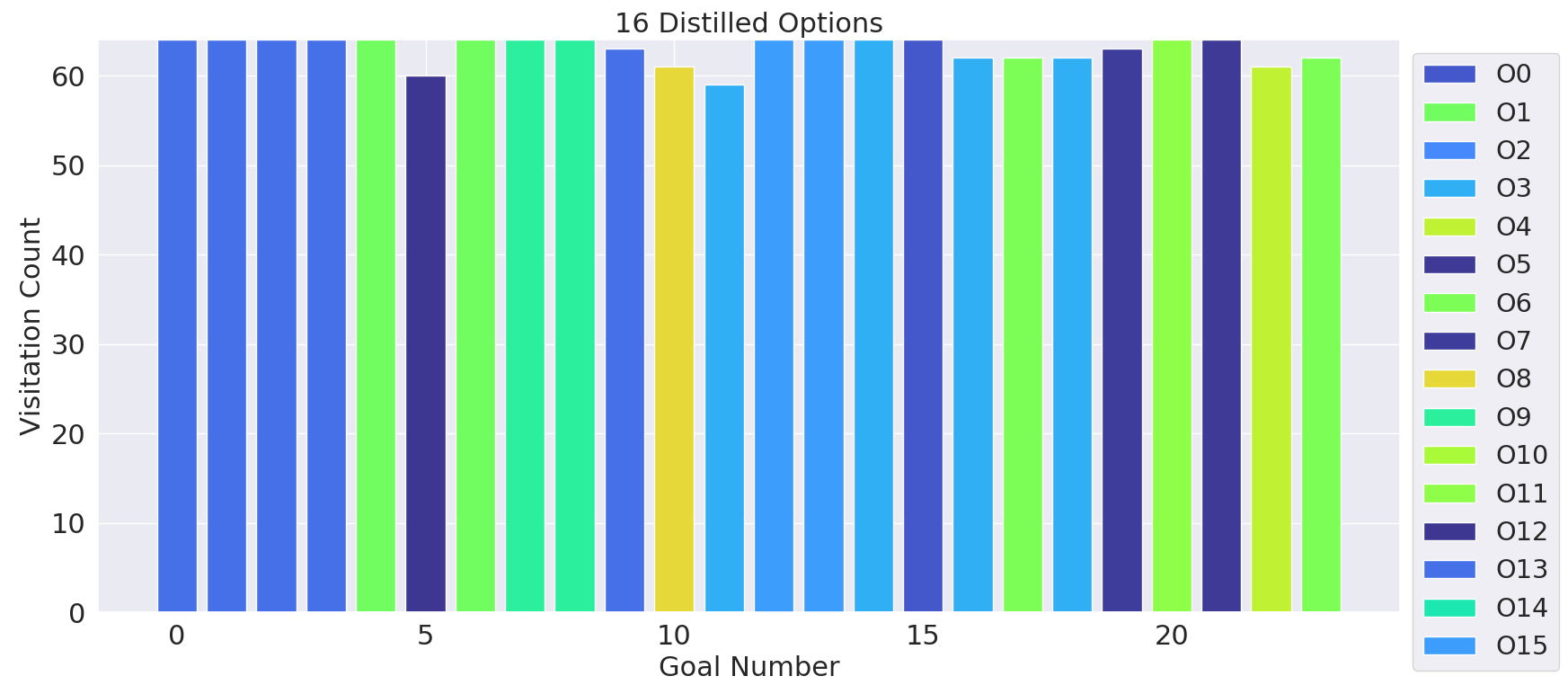}}
    \subfigure[AMN]{\label{fig:dmlabamn}\includegraphics[width=0.44\textwidth]{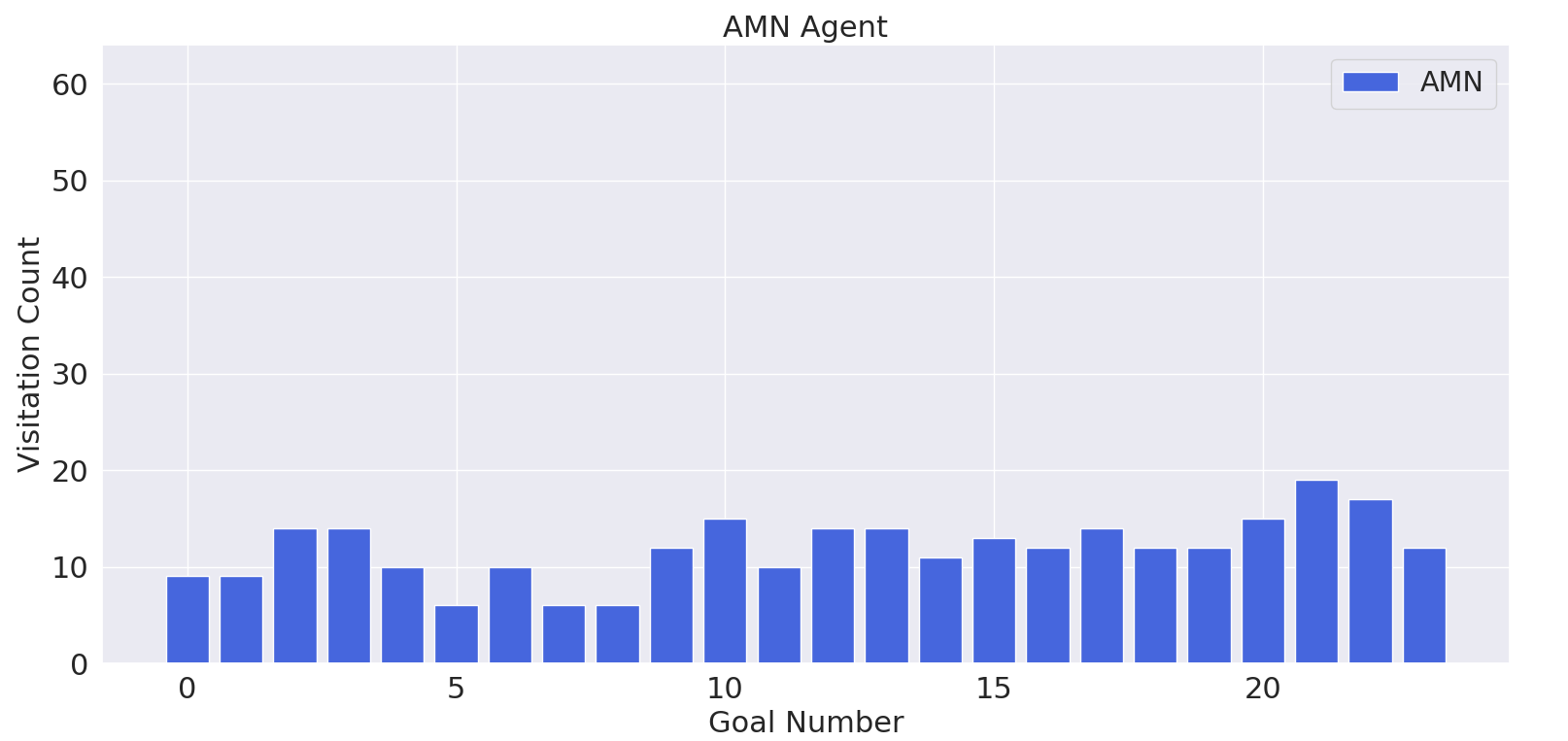}}
    \caption{Maximum visitation counts the distilled policies and AMN for the goal states}
\end{figure}

This section evaluates the Option Encoder on the Deepmind-lab maze domain task
depicted in Figure \ref{fig:dmenv}. We consider 24 option policies, trained
using PPO \cite{schulman2017proximal} to navigate to goal locations shown in
Figure \ref{fig:dmenv}. The agent starts from a random location. The expert
policies were distilled to 16,12 and 8 policies using the Option Encoder. Each
distilled policy was rolled out for 300 time steps and the visitation counts
were collected for all expert goal states. Figures
\ref{fig:dmlab8},\ref{fig:dmlab12},\ref{fig:dmlab16} depict the maximum value of
the visitation count among all the options. This is compared with AMN (see
Figure \ref{fig:dmlabamn}) where the option policies are distilled to a single
policy. The plots are indicative of the fact that the distilled policies
obtained from the Option Encoder cover a variety of goals and also visit them
more frequently. 
\section{Related Work}
Several works have attempted to address the policy distillation and compression
scenario. Prior works like Actor-mimic \cite{parisotto2015actor} have attempted
to compress a collection of policies. This framework can however distill a
collection of policies into a single policy. Our method, on the other hand, can
distill the expert policies into multiple basis policies. Pathnet
\cite{fernando2017pathnet} utilizes a large network, with weights being frozen
appropriately. However, like Actor-mimic, Pathnet only discovers a single policy
which may not have sufficient representational power. Pathnet addresses the
continual learning setup, where there is a sequence of tasks to be solved, which
may not be an appropriate approach to option pruning. The Elastic weight
consolidation loss \cite{kirkpatrick2017overcoming}, addresses a scenario
identical to that addressed in Pathnet \cite{fernando2017pathnet} and modifies
the weights using the gradient magnitude but suffers from same problems
described for the other two methods.

Option pruning has not been addressed in the context of option-discovery in
great detail. Option compression is not necessary for works like Option-critic
\cite{bacon2016optioncritic},  Deep Feudal reinforcement learning
\cite{vezhnevets2017feudal} or a collection of other works that discover options
relevant to the current task. On the other hand, task-agnostic option discovery
methods often need a large number of options to capture diverse behaviors.
McGovern et al. \cite{mcgovern2001automatic} use a collection of filters to
eliminate redundant options and \cite{machado2017laplacian} have observed an
improved performance when around 128 options are used in a rather modest 4-room
grid-world. Basis functions have been explored in the context of value functions
\cite{mahadevan2007proto,konidaris2011value}, where the structure of the graph
is used to define features for every state. On the other hand, our work uses a
set of options to define a basis over policies. Our work shares similarities
with the PG-ELLA \cite{ammar2014online} lifelong learning framework in that,
both attempt to discover a shared latent space from which a set of tasks are
solved. In this work, we focus on multi-task learning, where the basis policies
are utilized across a set of tasks.

\section{Conclusion}
In this work, we present the Option Encoder framework, which attempts to derive
a policy basis from a collection of option policies. The distilled policies can
be used as a substitute for the original set of options. We demonstrate the
utility of the distilled policies using an empirical evaluation on a collection
of tasks. As future work, one could extend the framework to work with value
functions. Another potential extension of this work is to the continual learning
framework, where the Option Encoder can be used to handle a set of new policies
and integrate the same with the policies learned earlier. This would involves
using the option-encoder in a batch-like manner, where the set of basis policies
are periodically refined. 

\bibliographystyle{splncs04}
\bibliography{ecml}

\begin{thebibliography}{10}
\providecommand{\url}[1]{\texttt{#1}}
\providecommand{\urlprefix}{URL }
\providecommand{\doi}[1]{https://doi.org/#1}

\bibitem{ammar2014online}
Ammar, H.B., Eaton, E., Ruvolo, P., Taylor, M.: Online multi-task learning for
  policy gradient methods. In: International Conference on Machine Learning.
  pp. 1206--1214 (2014)

\bibitem{bacon2016optioncritic}
Bacon, P.L., Harb, J., Precup, D.: The option-critic architecture. In:
  Thirty-First AAAI Conference on Artificial Intelligence (2017)

\bibitem{barreto2019option}
Barreto, A., Borsa, D., Hou, S., Comanici, G., Ayg{\"u}n, E., Hamel, P.,
  Toyama, D., Mourad, S., Silver, D., Precup, D., et~al.: The option keyboard:
  Combining skills in reinforcement learning. In: Advances in Neural
  Information Processing Systems. pp. 13031--13041 (2019)

\bibitem{beattie2016deepmind}
Beattie, C., Leibo, J.Z., Teplyashin, D., Ward, T., Wainwright, M.,
  K{\"u}ttler, H., Lefrancq, A., Green, S., Vald{\'e}s, V., Sadik, A., et~al.:
  Deepmind lab. arXiv preprint arXiv:1612.03801  (2016)

\bibitem{eysenbach2018diversity}
Eysenbach, B., Gupta, A., Ibarz, J., Levine, S.: Diversity is all you need:
  Learning skills without a reward function. arXiv preprint arXiv:1802.06070
  (2018)

\bibitem{fernando2017pathnet}
Fernando, C., Banarse, D., Blundell, C., Zwols, Y., Ha, D., Rusu, A.A.,
  Pritzel, A., Wierstra, D.: Pathnet: Evolution channels gradient descent in
  super neural networks. arXiv preprint arXiv:1701.08734  (2017)

\bibitem{kirkpatrick2017overcoming}
Kirkpatrick, J., Pascanu, R., Rabinowitz, N., Veness, J., Desjardins, G., Rusu,
  A.A., Milan, K., Quan, J., Ramalho, T., Grabska-Barwinska, A., et~al.:
  Overcoming catastrophic forgetting in neural networks. Proceedings of the
  national academy of sciences  \textbf{114}(13),  3521--3526 (2017)

\bibitem{konda2000actor}
Konda, V.R., Tsitsiklis, J.N.: Actor-critic algorithms. In: Advances in neural
  information processing systems. pp. 1008--1014 (2000)

\bibitem{konidaris2011value}
Konidaris, G., Osentoski, S., Thomas, P.: Value function approximation in
  reinforcement learning using the fourier basis. In: Twenty-fifth AAAI
  conference on artificial intelligence (2011)

\bibitem{lillicrap2015continuous}
Lillicrap, T.P., Hunt, J.J., Pritzel, A., Heess, N., Erez, T., Tassa, Y.,
  Silver, D., Wierstra, D.: Continuous control with deep reinforcement
  learning. arXiv preprint arXiv:1509.02971  (2015)

\bibitem{machado2017laplacian}
Machado, M.C., Bellemare, M.G., Bowling, M.: A laplacian framework for option
  discovery in reinforcement learning. arXiv preprint arXiv:1703.00956  (2017)

\bibitem{mahadevan2007proto}
Mahadevan, S., Maggioni, M.: Proto-value functions: A laplacian framework for
  learning representation and control in markov decision processes. Journal of
  Machine Learning Research  \textbf{8}(Oct),  2169--2231 (2007)

\bibitem{mcgovern2001automatic}
McGovern, A., Barto, A.G.: Automatic discovery of subgoals in reinforcement
  learning using diverse density  (2001)

\bibitem{menache2002q}
Menache, I., Mannor, S., Shimkin, N.: Q-cut—dynamic discovery of sub-goals in
  reinforcement learning. In: European Conference on Machine Learning. pp.
  295--306. Springer (2002)

\bibitem{mnih2016asynchronous}
Mnih, V., Badia, A.P., Mirza, M., Graves, A., Lillicrap, T., Harley, T.,
  Silver, D., Kavukcuoglu, K.: Asynchronous methods for deep reinforcement
  learning. In: International conference on machine learning. pp. 1928--1937
  (2016)

\bibitem{mnih2015humanlevel}
Mnih, V., Kavukcuoglu, K., Silver, D., Rusu, A.A., Veness, J., Bellemare, M.G.,
  Graves, A., Riedmiller, M., Fidjeland, A.K., Ostrovski, G., {others}:
  Human-level control through deep reinforcement learning. Nature
  \textbf{518}(7540), ~529 (2015)

\bibitem{parisotto2015actor}
Parisotto, E., Ba, J.L., Salakhutdinov, R.: Actor-mimic: Deep multitask and
  transfer reinforcement learning. arXiv preprint arXiv:1511.06342  (2015)

\bibitem{puterman1994markov}
Puterman, M.L.: Markov decision processes: Discrete stochastic dynamic
  programming  (1994)

\bibitem{rajendran2015attend}
Rajendran, J., Lakshminarayanan, A.S., Khapra, M.M., Prasanna, P., Ravindran,
  B.: Attend, adapt and transfer: Attentive deep architecture for adaptive
  transfer from multiple sources in the same domain. arXiv preprint
  arXiv:1510.02879  (2015)

\bibitem{schulman2017proximal}
Schulman, J., Wolski, F., Dhariwal, P., Radford, A., Klimov, O.: Proximal
  policy optimization algorithms. arXiv preprint arXiv:1707.06347  (2017)

\bibitem{silver2016mastering}
Silver, D., Huang, A., Maddison, C.J., Guez, A., Sifre, L., {van den
  Driessche}, G., Schrittwieser, J., Antonoglou, I., Panneershelvam, V.,
  Lanctot, M., Dieleman, S., Grewe, D., Nham, J., Kalchbrenner, N., Sutskever,
  I., Lillicrap, T., Leach, M., Kavukcuoglu, K., Graepel, T., Hassabis, D.:
  Mastering the game of {{Go}} with deep neural networks and tree search.
  Nature  \textbf{529}(7587),  484--489 (Jan 2016). \doi{10.1038/nature16961}

\bibitem{csimcsek2004using}
{\c{S}}im{\c{s}}ek, {\"O}., Barto, A.G.: Using relative novelty to identify
  useful temporal abstractions in reinforcement learning. In: Proceedings of
  the twenty-first international conference on Machine learning. p.~95. ACM
  (2004)

\bibitem{csimcsek2009skill}
{\c{S}}im{\c{s}}ek, {\"O}., Barto, A.G.: Skill characterization based on
  betweenness. In: Advances in neural information processing systems. pp.
  1497--1504 (2009)

\bibitem{simsek2005identifying}
{\c S}im{\c s}ek, O., Wolfe, A.P., Barto, A.G.: Identifying useful subgoals in
  reinforcement learning by local graph partitioning. pp. 816--823. {ACM Press}
  (2005). \doi{10.1145/1102351.1102454}

\bibitem{sutton1998reinforcement}
Sutton, R.S., Barto, A.G.: Reinforcement learning: An introduction. MIT press
  (1998)

\bibitem{sutton1999mdps}
Sutton, R.S., Precup, D., Singh, S.: Between {{MDPs}} and semi-{{MDPs}}: {{A}}
  framework for temporal abstraction in reinforcement learning. Artificial
  intelligence  \textbf{112}(1-2),  181--211 (1999)

\bibitem{vezhnevets2017feudal}
Vezhnevets, A.S., Osindero, S., Schaul, T., Heess, N., Jaderberg, M., Silver,
  D., Kavukcuoglu, K.: {{FeUdal Networks}} for {{Hierarchical Reinforcement
  Learning}}. arXiv:1703.01161 [cs]  (Mar 2017)

\end{thebibliography}

\clearpage
\end{document}